# Robust discovery of partial differential equations in complex situations

Hao Xu[a] and Dongxiao Zhang[b,c,*]

[a] *BIC-ESAT, ERE, and SKLTCS, College of Engineering, Peking University, Beijing 100871, P. R. China*

[b] *School of Environmental Science and Engineering, Southern University of Science and Technology, Shenzhen 518055, P. R. China*

[c] *Intelligent Energy Lab, Peng Cheng Laboratory, Shenzhen 518000, P. R. China*

[*] Corresponding author.

E-mail address: 390260267@pku.edu.cn (H. Xu); zhangdx@sustech.edu.cn (D. Zhang).

**Abstract**: Data-driven discovery of partial differential equations (PDEs) has achieved considerable development in recent years. Several aspects of problems have been resolved by sparse regression-based and neural network-based methods. However, the performances of existing methods lack stability when dealing with complex situations, including sparse data with high noise, high-order derivatives and shock waves, which bring obstacles to calculating derivatives accurately. Therefore, a robust PDE discovery framework, called the robust deep learning-genetic algorithm (R-DLGA), that incorporates the physics-informed neural network (PINN), is proposed in this work. In the framework, a preliminary result of potential terms provided by the deep learning-genetic algorithm is added into the loss function of the PINN as physical constraints to improve the accuracy of derivative calculation. It assists to optimize the preliminary result and obtain the ultimately discovered PDE by eliminating the error compensation terms. The stability and accuracy of the proposed R-DLGA in several complex situations are examined for proof-and-concept, and the results prove that the proposed framework is able to calculate derivatives accurately with the optimization of PINN and possesses surprising robustness to complex situations, including sparse data with high noise, high-order derivatives, and shock waves.

**Subject Areas:** Computational Physics, Machine Learning, PDE Discovery

## I. INTRODUCTION

Recently, data-driven discovery of partial differential equations (PDEs) has been studied in depth to explore additional possibilities for discovering underlying governing equations from available data. Sparse regression-based techniques, including Lasso [1], sequential threshold ridge regression (STRidge) [2], sparse Bayesian [3], SINDY [4] and other methods evolving from them [5–8], provide a paradigm for PDE discovery by identifying sparse PDE terms from a pre-determined candidate library which contains several potential terms. Meanwhile, deep-learning techniques, including physics-informed neural network (PINN) [9,10], PDE-network (PDE-NET) [11] and deep learning-PDE (DL-PDE) [12], are employed to improve the accuracy of derivative calculation via automatic differentiation of the neural network [13], which is crucial to the success of PDE discovery with noisy or sparse data. Unfortunately, although the above-mentioned methods work well with a small but complete candidate library, they are not sufficiently flexible to handle PDEs with high-order derivatives or unusual terms, which require the candidate library to contain numerous terms in order to remain complete. Therefore, genetic algorithm-based methods become a superior choice for PDE discovery since they are able to generate a large variation library via cross-over and mutation from a few basic genes [14–17]. Based on these fundamental techniques, more difficult problems are further solved to more closely approach practical applications. For example, parametric PDEs are successfully discovered by Xu et al. [18] and Rudy et al. [19] through stepwise deep learning-genetic algorithm (Stepwise DLGA) and sequential grouped threshold ridge regression (SGTR), respectively. Discovery of stochastic PDEs is accomplished by Bakarji and Tartakovsky [20] through transforming PDEs to probability density functions (PDFs). However, the performances of extant methods lack stability when dealing with complex situations, including sparse data with high noise, high-order derivatives, and PDE with shock waves.

Practically, experimental data are often precious and scarce, and may contain a certain level of noise due to inevitable systematic errors in the experiment, which may pose a challenge to PDE discovery. Although some of the above-mentioned methods have achieved satisfactory performances faced with a high level of noise, such as 25% noise [13,14] or even 50% in some cases [21], a large amount of data points is required to cope with high noise. Meanwhile, discovery of PDEs from sparse data also requires relatively clean data with low noise. In other words, it is difficult to handle sparse data with high noise. Another problem is the identification of PDEs with high-order derivatives, such as the Korteweg-de Vries (KdV) equation and the Kuramoto-Sivashinsky (KS) equation, since the accuracy of calculation of high-order derivatives decreases rapidly with the increase of noise level. Furthermore, a large amount of data is necessary to maintain the precision of derivative calculation. In order to handle this problem, recent works have attempted to discover the integral formation of PDEs as an alternative approach, which is able to decrease the required derivative order [22,23]. However, it is challenging to deal with PDEs that cannot be converted to integral form, and the integration process is time-intensive. The final challenge is PDE with shock waves, which exists in many physical processes, especially in the field of gas and fluid dynamics. At the position of the shock wave, the physical quantities change dramatically, which leads to a great deviation in derivatives calculation, and finally incurs failure in PDE discovery.

All of the challenges mentioned above are concentrated on one core issue: derivative calculation. As consequence, determination of how to maintain the accuracy of derivative calculation in complex conditions has become a vital problem because merely relying on finite difference or automatic differentiation is not sufficiently accurate to handle diversified situations. Regarding this issue, Both et al. [24] points out a promising approach to increase the accuracy of derivative calculation by employing the physics-informed neural network (PINN) to further optimize gradients. In their work, the outcome of Lasso is added into the loss function of PINN as physical constraints to improve the process of optimization, and results demonstrate that the derivative calculation is more accurate and more robust to sparse and noisy data. However, this method is unstable when facing high noise because the outcome is sensitive to the selection of the threshold which is employed to obtain sparsity. In addition, the candidate library in this work only contains 12 terms because it is difficult to maintain sparsity of the outcome when the size of candidate library is large. In other words, although the proposed method is able to improve the accuracy of derivative calculation, it is not stable enough for practical utilization. Therefore, it is essential to search for a robust method to discover PDEs in complex situations.

To deal with the problem of robustness and accuracy in complex situations, a method called the robust deep learning-genetic algorithm (R-DLGA) is proposed, which employs the deep learning-genetic algorithm (DLGA) to obtain a preliminary result of potential terms from a large variation library. It then utilizes these potential terms as physical constraints of PINN to further modify the derivatives, which assists to optimize the identified potential terms by eliminating the error compensation terms. Our proposed method does not require a pre-determined complete candidate library with numerous terms because additional potential terms can be augmented by DLGA. Consequently, it is easier to obtain sparsity in the process of PINN, and thereby increase the stability of the method. The proposed method is tested for proof-of-concept with several cases in complex situations that are difficult to be resolved by existing methods, including the KdV equation with sparse and noisy data, the KS equation with high-order derivatives, and the Burgers equation with shock waves. Satisfactory outcomes are obtained, and our proposed method maintains accuracy and robustness in these complex situations.

The remainder of this paper proceeds as follows. Section II establishes the framework of R-DLGA. Section III presents the performances of PDE discovery with our proposed algorithm in complex conditions, and illustrates the robustness and accuracy of R-DLGA. Finally, Section IV summarizes the advantages and disadvantages of the proposed algorithm, and provides remarks and recommendations for future work.

## II. METHODS

### A. Problem settings

For PDE discovery, our goal is to discover the form of underlying governing equations from spatial-temporal observation data, which is given as $u(x,t)$. The form of PDE can be written as:

$$u_t = F(u, u_x, uu_x, u^2 u_x, u_{xx}, \ldots; \vec{\xi}), \tag{1}$$

where $F$ is an operator that indicates a linear combination of terms; and $\vec{\xi}$ is the coefficient vector.

With available observation data $u(x_i, t_i)$, this issue can be converted to a regression problem that is described as follows:

$$\begin{bmatrix} u_t(x_1,t_1) \\ u_t(x_2,t_2) \\ \vdots \\ u_t(x_N,t_N) \end{bmatrix} = \begin{bmatrix} u(x_1,t_1) & \cdots & uu_x(x_1,t_1) & \cdots \\ u(x_2,t_2) & \cdots & uu_x(x_2,t_2) & \cdots \\ \vdots & \ddots & \vdots & \ddots \\ u(x_N,t_N) & \cdots & uu_x(x_N,t_N) & \cdots \end{bmatrix} \cdot \vec{\xi}, \qquad (2)$$

where $N$ is the data volume; and $\vec{\xi}$ is the coefficient vector. Eq. (2) can be abbreviated as:

$$U_t = \Phi \cdot \vec{\xi}. \qquad (3)$$

where $U_t$ refers to the left-hand side term; and $\Phi$ refers to the matrix of the potential terms. It is worth noting that $\Phi$ may contain a large variety of potential terms, and thus the coefficient vector $\xi$ is a sparse vector with most of the elements being 0. For PDE discovery, it aims to identify non-zero terms and corresponding coefficients from data. Therefore, the difficulty of the problem lies in how to give consideration to both accuracy and parsimony of the discovered PDE.

### B. Deep learning-genetic algorithm (DLGA)

In our earlier work, an algorithm combining deep-learning and genetic algorithm, called the deep learning-genetic algorithm (DLGA), is proposed to discover PDE [14]. Although it manages to discover many common PDEs, it is unstable in complex situations, as mentioned in Section I. In this work, a novel generalized genetic algorithm is proposed based on the DLGA framework to give a preliminary result of potential terms that is relatively parsimonious and contains certain physical information. The work flow of our proposed robust deep learning-genetic algorithm (R-DLGA) is demonstrated in Fig. 1. The DLGA steps are illustrated in Fig. 1(a) and the PINN steps are illustrated in Fig. 1(b). In the DLGA process, a neural network is employed to construct a substitute model from available observation data. Then, a large amount of meta-data is generated on a regular spatial-temporal grid, and automatic differentiation is utilized to calculate the derivatives of the meta-data. Finally, the generalized genetic algorithm is utilized to give a preliminary result of potential terms. In this part, the procedure of DLGA steps, including the neural network and generalized genetic algorithm, is introduced in detail.

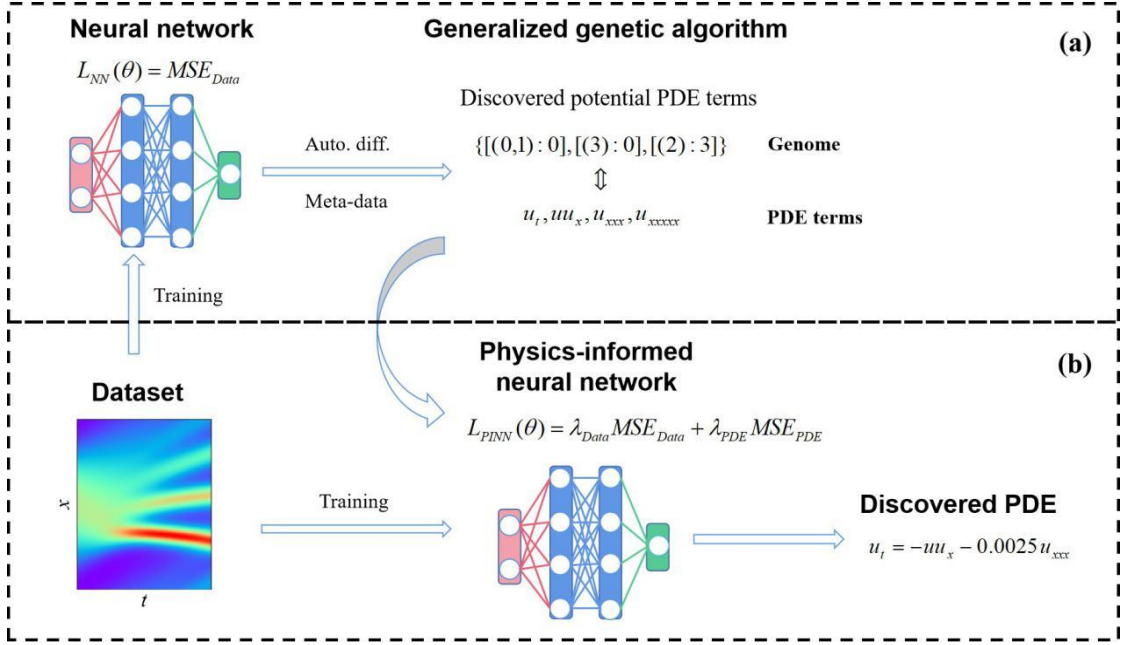

**FIG. 1.** Work flow of R-DLGA, including the DLGA steps (a) and the PINN steps (b). In the DLGA steps, the neural network is utilized to construct a surrogate model from available data, and then the generalized genetic algorithm is employed to identify potential terms. In the PINN steps, the discovered potential terms are added into the loss function $L_{PINN}(\theta)$ as physical constraints to further optimize the derivatives and discover the ultimate PDE.

*1. Neural network*

In this work, the neural network is a fully-connected artificial neural network (ANN), which is composed of an input layer, an output layer, and several hidden layers. The input of the neural network is the spatial-temporal location $(x,t)$, and the output is $NN(x,t;\theta)$. Here, $\theta$ is the parameter of the neural network, including weights and bias. The neural network is trained by minimizing the loss function $L_{NN}(\theta)$, which is written as:

$$L_{NN}(\theta) = MSE_{Data} = \frac{1}{N}\sum_{i=1}^{N}[u(x_i,t_i) - NN(x_i,t_i;\theta)]^2, \tag{4}$$

where $N$ is the data volume. An early termination technique is employed to prevent over-fitting. The trained neural network functions as a surrogate model for the underlying physical system, and is utilized to generate meta-data on a spatial-temporal grid. Meanwhile, automatic differentiation of the neural network is employed to calculate derivatives of meta-data, which has been proven to be more robust to data noise.

*2. Generalized genetic algorithm*

The genetic algorithm is an optimization algorithm that simulates the evolutionary process of natural selection, which is usually comprised of digitization, cross-over, mutation, fitness calculation, and evolution. In this work, a new genetic algorithm, called the generalized genetic algorithm, is proposed. Different from the common genetic algorithm, the generalized genetic algorithm attempts to consider PDE terms in compound form that consists of an inner term and a derivative order. In this way, PDE terms can be expressed flexibly in a general form, and high-order derivatives can be reached by order mutation without prior definition. Therefore, the error compensation terms with tiny coefficients that are composed of high-order derivatives can be discovered in complex situations, which compensates most of error resultant from derivative calculation and guarantees the stability and accuracy of the correct terms. In this part, the procedure of our proposed generalized genetic algorithm for PDE discovery is introduced in detail.

*(a) Digitization*

In this work, a unique digitization technique is put forward to express possible terms clearly and flexibly in general form. The principle of digitization is illustrated in Fig. 2. Different from previous works [14,17] that only consider the simple form of terms (e.g., $uu_x$), our proposed digitization technique manages to express terms in compound form (e.g., $(u^2)_{xx}$). Considering the existence of the weak form of PDEs, most PDE terms can be written in compound form, which is composed of an inner term and a derivative order. The relation between PDE terms in compound form and genetic form is illustrated in Fig. 2(a).

The gene is the basic unit of the genetic algorithm, in which numbers are employed to represent the derivatives of corresponding order. Here, basic genes are defined up to three-order derivatives to form the inner term, which is represented by parentheses. It is assumed that there is only multiplication of basic genes in the inner term. For example, (0,2) refers to the inner term $uu_{xx}$. With the inner term and the derivative order, the compound form of PDE terms can be established in square brackets, which is called the gene module. In the gene module, the inner term and the derivative order are separated by colons. Since each gene module represents a corresponding PDE term, the PDE can be expressed as the addition of multiple terms, which is called the genome. The genome is represented by curly brackets, and different gene modules are separated by commas. The whole digitization process from basic gene to genome is displayed in Fig. 2(b). It is worth mentioning that the meta-data generated by the trained neural network $NN(x,t;\theta)$ are utilized here, derivatives in basic genes are calculated by automatic differentiation, and the value of the inner term can be calculated easily by multiplication. Meanwhile, the value of the PDE term in compound form is calculated by the finite difference method since the value of the inner term is calculated on a regular grid. In this manner, the calculation of PDE terms in compound form, which is unsettled in previous works [14,17], is resolved in this work.

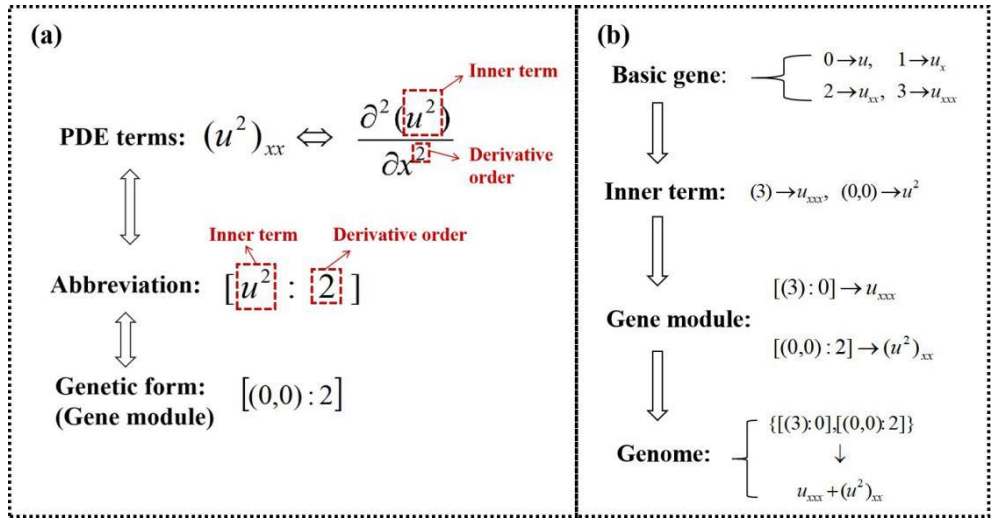

**FIG. 2.** The principle of digitization in the generalized genetic algorithm. (a) The relation between PDE terms in compound form and genetic form; (b) the digitization process from basic gene to genome.

*(b) Cross-over and mutation*

Cross-over and mutation are crucial to the genetic algorithm because they can generate new genomes and thereby expand the search scope, which leads to a large variation library of terms. A diagrammatic sketch of cross-over and mutation is provided in Fig. 3. In the cross-over process, two parent genomes exchange certain gene modules to produce their children, which enables the gene module in the parent genomes to transfer to the next generation. In the mutation process, the genome randomly mutates to a new genome. In this work, there are four ways of mutation, including delete module mutation, basic gene mutation, order mutation, and add module mutation. For delete module mutation, a gene module is randomly deleted. For basic gene mutation, one random gene in the gene module is mutated by decreasing 1 from the gene number. Particularly, 0 is mutated to be 3. For example, (1,0) may be mutated into (0,0), and (0,2) may be mutated into (3,2). For order mutation, the derivative order in a certain gene module is mutated in the same way as basic gene mutation. For add module mutation, a random gene module is added into the genome.

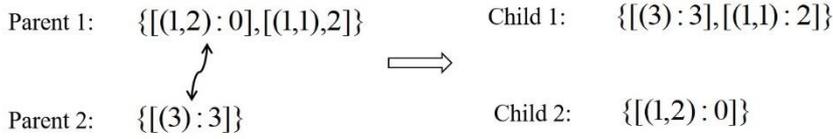

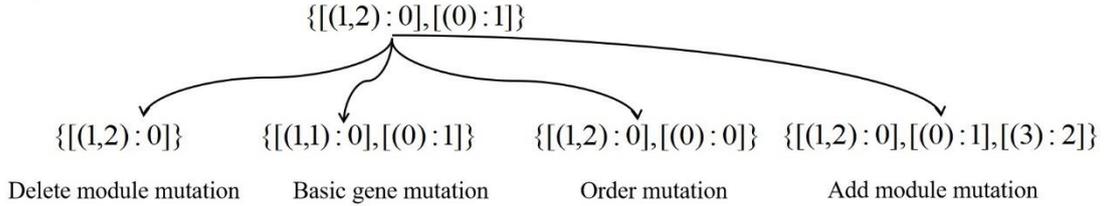

**FIG. 3.** The diagrammatic sketch of cross-over (a) and mutation (b) in the generalized genetic algorithm.

*(c) Fitness calculation*

In the genetic algorithm, the quality of the genome is determined by the fitness function, which is defined as:

$$F = MSE + \varepsilon \sum_{k=1}^{N_{GM}} L(GM_k), \qquad (5)$$

with

$$MSE = \frac{1}{N_x N_t} \sum_{j=1}^{N_t} \sum_{i=1}^{N_x} \left| U_t(x_i, t_j) - \vec{\beta} U_R(x_i, t_j) \right|^2, \qquad (6)$$

where $F$ denotes the fitness. The fitness contains two parts, including the $MSE$ part and the penalty part. $MSE$ is calculated according to Eq. (6), where $U_t(x_i, t_j)$ is the value of the left-hand term with the size of $1\times 1$; $U_R(x_i,t_j)$ is the value of right-hand terms translated from the genome, which is a $n\times 1$ vector; $n$ is the number of PDE terms; and $N_x$ and $N_t$ are the number of $x$ and $t$ of the meta-data, respectively. $\vec{\beta}$ with the size of $1\times n$ is the coefficient for the right-hand terms which is calculated by least square regression. In the penalty part, $\varepsilon$ is a hyper-parameter to control the weight of penalty. $N_{GM}$ denotes the number of gene modules in the genome, and $L(GM_k)$ refers to the length of $k^{th}$ gene module in the genome, i.e., the number of genes in the inner term. For example, the length of gene module [(1,2):0] is 2.

The *MSE* part in the fitness function is utilized to reflect the accuracy of the discovered PDE. Specifically, the smaller is the *MSE*, the more accurate is the discovered PDE. However, considering that a parsimonious form is expected to be discovered, a penalty is needed to be added into the fitness to avoid over-fitting. The penalty part in Eq. (5) limits the total length of the genome, which enables the discovered PDE to tend to be parsimonious. Therefore, in this work, the smaller is the fitness function, the better is the genome.

*(d) Evolution*

The work flow for the whole evolution process in the generalized genetic algorithm is presented in Fig. 4. In this work, genomes are randomly generated on the basis of basic genes as the initial parent generation. Then, each parent genome cross-overs twice to produce twice the number of children (e.g., 200 parents produce 400 children). Afterwards, four types of mutation occur independently, and the fitness of the children is calculated. The best half of children with smaller fitness in each generation is reserved to be the subsequent parent generation, while the others are dropped. This cycle continues until achieving maximum generations, and the best children in the final generation are the ultimate identified potential terms. A strategy is employed here to accelerate the convergence, in which the best child in each generation is fixed until a better one occurs to replace it.

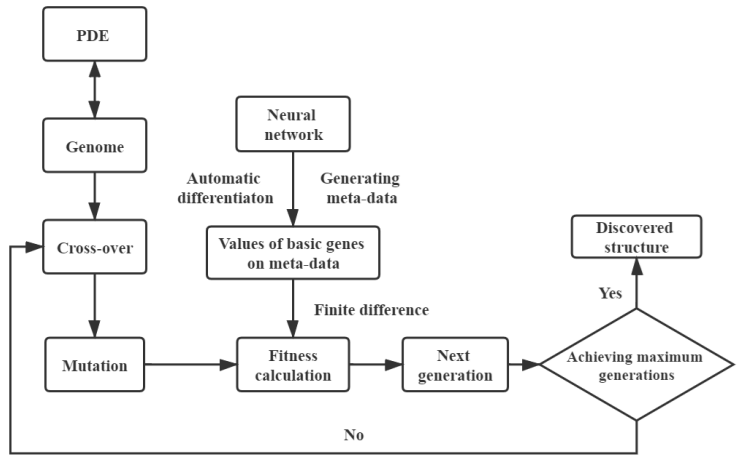

FIG. 4. The work flow for the whole evolution process in the generalized genetic algorithm.

**C. Physics-informed neural network (PINN)**

After the DLGA process, a preliminary result of potential terms is obtained, which may be imprecise in complex conditions, but still contains some physical information. Therefore, the identified potential terms can be added into the loss function of the neural network $NN(x,t;\theta)$ to provide physical constraints and construct a physics-informed neural network $PINN(x,t;\theta)$ [9,10]. The structure, the weights, and the bias of physics-informed neural network $PINN(x,t;\theta)$ are the same as those of the trained neural network $NN(x,t;\theta)$, which means that $PINN(x,t;\theta)$ is trained on the basis of $NN(x,t;\theta)$. The input is also the same spatial-temporal locations $(x,t)$, and the output is $PINN(x,t;\theta)$. The loss function of $PINN(x,t;\theta)$ is:

$$L_{NN_2}(\theta) = \lambda_{Data}MSE_{Data} + \lambda_{PDE}MSE_{PDE}$$
$$= \frac{\lambda_{Data}}{N}\sum_{i=1}^{N}(u(x_i,t_i) - NN(x_i,t_i))^2 + \frac{\lambda_{PDE}}{N}\sum_{i=1}^{N}\left|U_t(x_i,t_i) - \vec{\beta}_X U_X(x_i,t_i)\right|^2 \quad (7)$$

where $N$ is the number of observation data; $MSE_{data}$ is the same as that in Eq. (4), and $MSE_{PDE}$ is the mean squared error between the left-hand side term and the potential terms on the right-hand side. In this work, $\lambda_{Data}$ and $\lambda_{PDE}$, which are corresponding coefficients of the two constraint terms, are both set to be 1. It is worth noting that these coefficients can be adjusted according to the magnitude of $MSE_{Data}$ and $MSE_{PDE}$ for better outcomes. Furthermore, other constraint terms, such as boundary or initial conditions and engineering controls, may be included in Eq. (7) [25] if relevant prior information is known, and the weighting coefficients for these constraints may be determined automatically [26]. In Eq. (7), $U_t(x_i,t_i)$ is the value of the left-hand side term, and $U_X(x_i,t_i)$ is the value of preliminary potential terms identified by DLGA with the size of $n_p \times 1$, where $n_p$ is the number of potential terms and $\vec{\beta}_X$ with the size of $1 \times n_p$ is the coefficient for the right-hand terms. In each training epoch, $U_t(x_i,t_i)$ and $U_X(x_i,t_i)$ can be calculated easily via automatic differentiation. Therefore, the coefficient vector $\vec{\beta}_X$ can be obtained by regression techniques. For the first 1000 epochs, the Lasso regression technique [1] is employed, which is written as:

$$\vec{\beta}_X = \mathrm{argmin}_{\vec{\beta}}(\left\|U_t - \vec{\beta}U_X\right\|_2^2 + \alpha\left\|\vec{\beta}\right\|_1), \quad (8)$$

where $\alpha$ is the weight of $L_1$ normalization and is set to be $10^{-4}$ in this work. It is worth noting that $L_1$ normalization is utilized in Lasso to further shrink the coefficient vector $\vec{\beta}_X$ in order to better distinguish the correct terms from error compensation terms. When the training epoch>1000, the term whose absolute value of coefficient is smaller than a threshold $\lambda$ will be dropped in each epoch. The value of the threshold $\lambda$ is fixed to be $10^{-4}$ in this work. It is worth noting that in previous work [24], the choice of threshold is crucial and must be changed frequently in order to discover different PDEs in different situations, especially when faced with high levels of noise, which means that a sophisticated procedure is needed to adjust the threshold. In comparison, in this work, the threshold is fixed, which makes our proposed method more robust. Moreover, considering that Lasso will bring errors to the calculated coefficients, least square regression is employed for the rest of the epochs, which is written as:

$$\vec{\beta}_X = (U_X^T U_X)^{-1} U_X^T U_t. \quad (9)$$

When the training epoch approaches the maximum epoch, the training process will be stopped,

and the ultimate non-zero terms are the discovered terms and $\vec{\beta}_X$ is the corresponding coefficients. It has been proven in previous work [24] that the correct physical constraint can assist to improve the derivative calculation of the neural network. In this work, although the preliminary discovered potential terms by the generalized genetic algorithm may be incorrect, they still contain underlying information and are close to the true physical process. Therefore, we construct an optimization cycle by PINN, which is illustrated in Fig. 5. In the optimization cycle, the physical constraints (i.e., the discovered potential terms) make the neural network closer to the true physical process, and thereby improve the accuracy of derivative calculation, while the improvement of derivative calculation will make the physical constraints more accurate. Overall, $PINN(x,t;\theta)$ utilizes the underlying physical information contained in the preliminary result of potential terms to further optimize the structure of the neural network to get closer to the actual physical processes, which improves the accuracy and stability of the discovered PDE.

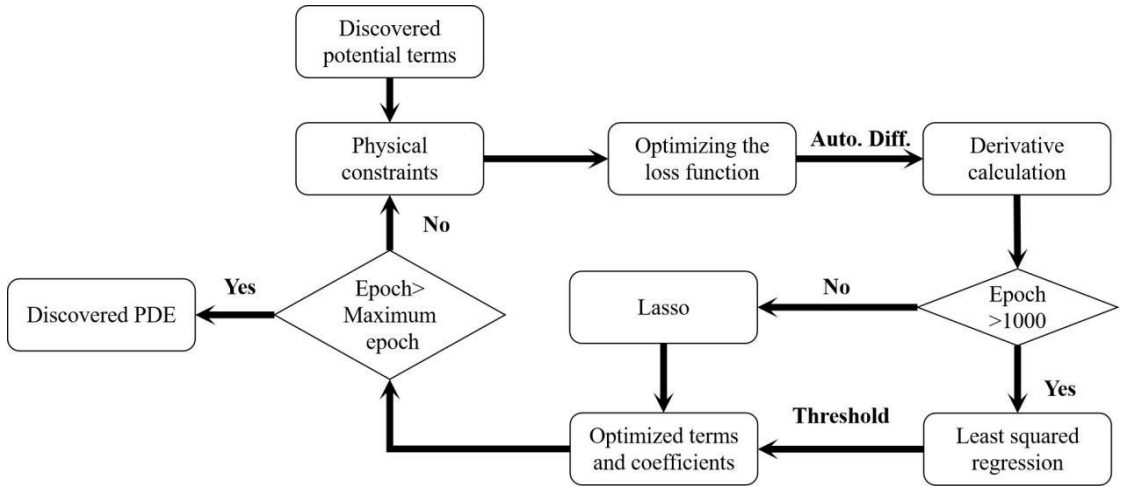

**FIG. 5.** The work flow for the optimization cycle in PINN. Auto. Diff. refers to automatic differentiation.

## III. RESULTS

In this section, the performance of our proposed R-DLGA in three typical complex situations is tested, including sparse data with high noise, high-order derivatives, and shock waves. The KdV equation, the KS equation, and the Burgers equation are taken as examples. These examples are employed to examine the robustness and accuracy of R-DLGA when dealing with various difficult cases. The datasets utilized in this work are provided as an open resource in the website https://gitee.com/xh251314/R_DLGA/tree/master.

In this work, two neural networks $NN(x,t;\theta)$ and $PINN(x,t;\theta)$ both have five layers, an input layer, an output layer and three hidden layers, with 50 neurons in each hidden layer. The maximum training epoch for $NN(x,t;\theta)$ and $PINN(x,t;\theta)$ is 50,000 and 20,000, respectively. The activation function is $sin(x)$. The value of threshold is $10^{-4}$. For the generalized genetic algorithm, the population size of genomes is 400, and the number of maximum generations is 100. The rate of cross-over is 0.8, the rate of order mutation and basic gene mutation is 0.3, the rate of delete module mutation is 0.5, and the rate of add module mutation is 0.4. $l_1$_norm in Lasso is $10^{-4}$. The hyper-parameter $\varepsilon$ is 0.1 for the KdV equation, and 0.01 for the KS equation and the Burgers equation.

### A. Discovery of KdV equation from sparse data with high noise

The KdV equation is a partial differential equation discovered by Korteweg and de Vries [27] to describe the motion of unidirectional shallow water, which is expressed as:

$$u_t = -uu_x - 0.0025u_{xxx}. \tag{10}$$

The dataset is generated by numerical simulation with 512 spatial observation points in the domain $x \in [-1,1)$ and 201 temporal observation points in the domain $t \in [0,1]$. Therefore, the data volume is 102,912. For meta-data, there are 400 spatial points in the domain $x \in [-0.8, 0.8]$ and 300 temporal observation points in the domain $t \in [0.1, 0.9]$, and thus the total number of meta-data is 120,000. In order to better demonstrate the stability and accuracy of R-DLGA facing sparse and noisy data, 25,000 (24.4% of the total data volume), 10,000 (9.80%), 2,500 (2.44%), 1,000 (1.22%), 500 (0.44%), and 100 (0.09%) data are randomly selected to form new datasets. 0% noise (clean data), 5% noise, 15% noise, 25% noise, and 50% noise are added to the dataset in the following form:

$$u(x,t) = u(x,t) \cdot (1 + \gamma \times e), \tag{11}$$

where $\gamma$ denotes the noise level; and $e$ is a uniform random variable, taking values from -1 to 1. The result is shown in Fig. 6 (a). Here, the relative error is calculated in the following way:

$$error = \left( \frac{\sum_{j=1}^{N_t} \sum_{i=1}^{N_x} |u(x_i, t_j) - u'(x_i, t_j)|^2}{\sum_{j=1}^{N_t} \sum_{i=1}^{N_x} |u(x_i, t_j)|^2} \right)^{\frac{1}{2}} \times 100\%, \tag{12}$$

where $u(x_i, t_j)$ is the solution of the correct PDE; $u'(x_i, t_j)$ is the solution of the discovered PDE; and $N_x$ and $N_t$ are the number of $x$ and $t$, respectively. Meanwhile, the discovered PDEs from these new noisy datasets by DLGA alone are illustrated in Fig. 6(b) for comparison. From the figure, it is observed that, although DLGA alone is able to discover PDEs with relatively high levels of noise (15% noise), the error is large. In contrast, R-DLGA is surprisingly robust to small data volume and high noise. The outcome remains stable and accurate even with 500 data (0.49%) and 50% noise. Moreover, under the same conditions, the error of PDE discovered by R-DLGA is much smaller, especially when the noise is high and the amount of data is small. This demonstrates that the physical constraints in the PINN process provided by the preliminary result of potential terms are able to improve the performance of derivative calculation, which enables our proposed algorithm to be more robust to small data volume with large data noise.

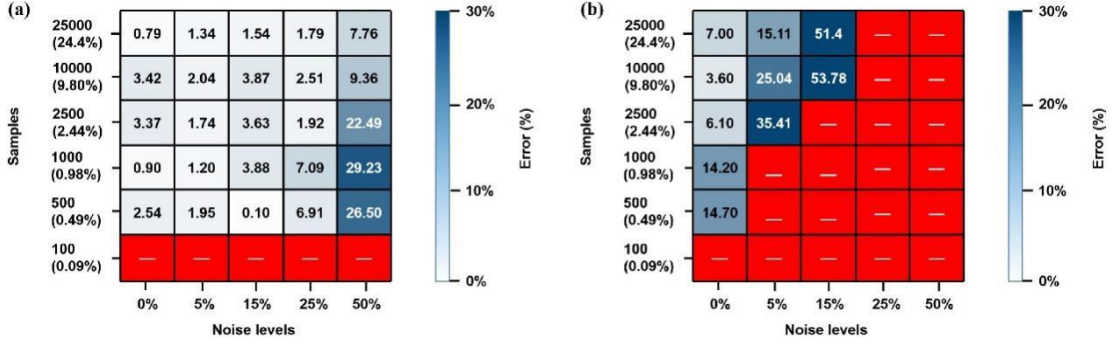

**FIG. 6.** The identified KdV equation with different volumes of data and different levels of noise through R-DLGA (a) and DLGA alone (b). The red grid indicates that an incorrect PDE is discovered, while the blue grid indicates that a correct PDE is discovered. The value in the grid refers to the relative error. Here, the darker is the blue color, the larger is the relative error.

### B. Discovery of KS equation with high-order derivatives

In this part, the ability of R-DLGA to discover PDEs with high-order derivatives is tested with the KS equation, the form of which is expressed as:

$$u_t = -uu_x - u_{xx} - u_{xxxx}. \tag{13}$$

The dataset is generated by numerical simulation with 512 spatial observation points in $x \in [-10,10)$ and 251 temporal observation points in $t \in [0,50]$. As a result, the total number of data is 128,512. For meta-data, there are 400 spatial points in $x \in [-6,6]$ and 300 temporal observation points in $t \in [10,40]$, and thus the total number of meta-data is 120,000. The KS equation has a fourth-order derivative which is difficult to calculate accurately through existing derivative calculation methods, especially with noisy data. In this experiment, 60,000 data are randomly selected to train the neural network, and 0% noise (clean data), 1% noise, 5% noise, and 10% noise are added to the data. The result is shown in Table I.

**TABLE I**. The KS equation identified by R-DLGA with different levels of noise added to the data.

| Correct PDE: $u_t = -uu_x - u_{xx} - u_{xxxx}$ | | |
|---|---|---|
| **Noise Level** | **Learned Equation** | **Error** |
| Clean data | $u_t = -0.999uu_x - 1.000u_{xx} - 1.000u_{xxxx}$ | 2.93% |
| 1% noise | $u_t = -0.995uu_x - 0.994u_{xx} - 0.994u_{xxxx}$ | 3.16% |
| 5% noise | $u_t = -0.992uu_x - 0.992u_{xx} - 0.992u_{xxxx}$ | 6.12% |
| 10% noise | $u_t = -0.325uu_x - 0.288u_{xx} - 0.195u$ | -- |

Until now, the best result is obtained by the integral form [28], which decreases the derivative order to the third-order, which is easier to be calculated and is robust to 5% noise. However, although the PDE form can be discovered, the accuracy is still unsatisfactory, with 25% relative error for clean data and 95% relative error for 5% noise. In comparison, the error of PDE discovered by R-DLGA is much smaller, with 2.93% relative error for clean data and 6.12% relative error for 5% noise. This indicates that R-DLGA is able to calculate high-order derivatives accurately by adding the preliminary result of potential terms to the loss function of PINN as physical constraints.

## C. Discovery of Burgers equation with shock waves

The Burgers equation has been identified many times for proof-of-concept in previous works [2,10], and its form is written as:

$$u_t = -uu_x + au_{xx} \tag{14}$$

where $a$ is the coefficient of the viscous term. In previous investigations [2,10], $a$ is set to be 0.1, and the solution is smooth without a shock wave. In this work, a more challenging situation is considered in which $a$ is set to be $\frac{0.01}{\pi}$, which means that the viscous term is so small that a shock wave will emerge. The dataset is the same as that in Both et al. [24]. For the dataset, there are 256 spatial observation points in $x \in [-1,1]$ and 100 temporal observation points in $t \in [0,1)$. Consequently, the total number of data is 25,600. For meta-data, there are 400 spatial points in $x \in [-0.8, 0.8]$ and 300 temporal observation points in $t \in [0.1, 0.9]$, and thus the total number of meta-data is 120,000. 5,000 data (19.53% of the total data volume) are randomly selected to train the neural network, and 0% noise (clean data) and 25% noise are added to the data. The results are displayed in Fig. 7(a) and (b), respectively. From the figure, it can be seen that the discovered PDE is close to the true PDE, and the shock wave is recovered with high accuracy, which means that R-DLGA is able to handle PDE with shock waves even though the derivative calculation is difficult at the location of the shock wave. In addition, it is found that the performance of R-DLGA when identifying the Burgers equation with a shock wave is insensitive to data noise.

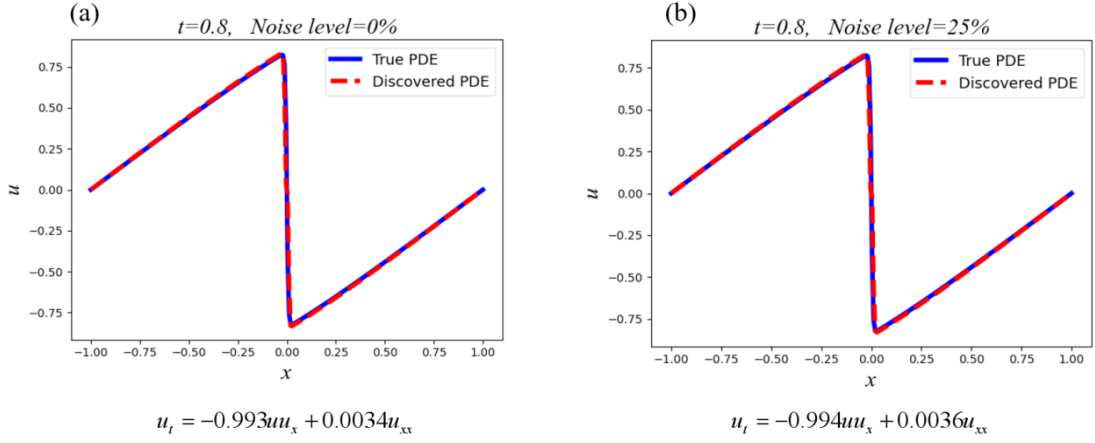

$$u_t = -0.993uu_x + 0.0034u_{xx} \qquad u_t = -0.994uu_x + 0.0036u_{xx}$$

**FIG. 7.** The solution of identified PDE and true PDE for the Burgers equation with a shock wave when $t$=0.8 with clean data (a) and 25% noise (b).

### D. Effect of the generalized genetic algorithm

In this work, a unique genetic algorithm, called the generalized genetic algorithm, is proposed to obtain a preliminary result of potential terms, in which terms are expressed in the compound form comprised of an inner term and a derivative order. With the generalized genetic algorithm, an interesting phenomenon is discovered. The KdV equation with 2,500 data training the neural network is taken as an example again, and different levels of noise are added to the data. The results of the generalized genetic algorithm, including discovered potential term and corresponding coefficients, are presented in Table II. For comparison, the ultimate discovered PDE by R-DLGA is also provided in Table II. From the table, it can be seen that the identified potential PDE terms are correct when the noise level is low, while redundant potential terms are found when the noise level is high. However, it is surprising to find that several high-order terms with tiny coefficients (e.g., $u_{xxxxx}$ and $(u^2)_{xxx}$) occur in the potential terms, while the correct terms are also in the potential terms with their coefficients being relatively accurate. This means that these high-order terms function as error compensation terms to keep the coefficients of correct terms relatively accurate, which guarantees the stability of the identification of the correct terms in the potential terms. Meanwhile, the error compensation terms will be dropped with the optimization of derivative calculation during the PINN steps.

This phenomenon can be discovered because the generalized genetic algorithm adds the derivative order into the gene module and attempts to find a compound form which is more concise and flexible, so that the genome can automatically increase the derivative order via mutation to identify a better solution. In contrast, previous methods usually consider up to fourth-order derivative at most, and only a simple form can be generated, and thus it is difficult to find suitable error compensation terms. To better illustrate the superiority of our proposed generalized genetic algorithm, different PDE discovery methods, including STRidge, common genetic algorithm, generalized genetic algorithm and R-DLGA, are adopted to discover the KdV equation with 2,500 data and 50% noise. The results are presented in Table III. It can be found that only R-DLGA discovered the correct PDE and the coefficients are accurate, while the others fail. Among them, STRidge is completely incorrect and, although the true terms are also contained in the PDE discovered by the common genetic algorithm, their coefficients are far from the correct coefficients. This is because it fails to discover correct error compensation terms. In contrast, the generalized genetic algorithm is able to discover the suitable error compensation terms, and the coefficients of the correct term are relatively accurate.

TABLE II. Potential terms and corresponding coefficients discovered by the generalized genetic algorithm and ultimately learned PDE by R-DLGA with different levels of noise added to the data for discovering the KdV equation.

| | Correct PDE: $u_t = -0.0025u_{xxx} - uu_x$ | |
|---|---|---|
| Noise Level | Learned Equation by Generalized Genetic Algorithm | Learned Equation by R-DLGA |
| Clean data | $u_t = -0.00247u_{xxx} - 0.494(u^2)_x$ | $u_t = -0.00244u_{xxx} - 0.954uu_x$ |
| 5% noise | $u_t = 0.00231u_{xxx} - 0.466(u^2)_x$ | $u_t = -0.00250u_{xxx} - 0.997uu_x$ |
| 15% noise | $u_t = -0.00230u_{xxx} - 0.447(u^2)_x - 2.11 \times 10^{-7}(u_{xx})_{xxx}$ | $u_t = -0.00249u_{xxx} - 0.993uu_x$ |
| 25% noise | $u_t = -0.00247u_{xxx} - 0.994uu_x - 8.67 \times 10^{-7}(u_{xx})_{xxx} - 1.68 \times 10^{-4}(u^2)_{xxx}$ | $u_t = -0.00250u_{xxx} - 1.002uu_x$ |
| 50% noise | $u_t = -0.00225u_{xxx} - 0.895uu_x - 9.92 \times 10^{-7}(u_{xx})_{xxx} - 1.86 \times 10^{-4}(u^2)_{xxx}$ | $u_t = -0.00249u_{xxx} - 0.993uu_x$ |

TABLE III. PDE discovered by different methods for the KdV equation from sparse data with high noise.

| | Correct PDE: $u_t = -0.0025u_{xxx} - uu_x$ |
|---|---|
| Learned equation by STRidge | $u_t = -0.807uu_x$ |
| Learned equation by common genetic algorithm | $u_t = -0.00129u_{xxx} - 0.447uu_x - 3.01 \times 10^{-4}u_x^3$ |
| Learned equation by generalized genetic algorithm | $u_t = -0.00247u_{xxx} - 0.994uu_x - 8.67 \times 10^{-7}(u_{xx})_{xxx} - 1.68 \times 10^{-4}(u^2)_{xxx}$ |
| Learned equation by R-DLGA | $u_t = -0.00250u_{xxx} - 1.002uu_x$ |

### E. The effect of PINN steps

In this part, the effect of PINN steps is investigated. The KdV equation with 500 data and 50% noise at $t=0.4$, and the Burgers equation with 5,000 data and 25% noise at $t=0.8$, are examined again. True data, noisy data, and the outcome of $NN(x,t;\theta)$ and $PINN(x,t;\theta)$ are plotted in Fig. 8. From the figure, numerous findings are evident. Firstly, with a high level of noise, the noisy observation data have a large derivation compared with the true data, and the neural network is able to learn the underlying physical dynamics from the sparse data with high noise. This indicates that the DLGA steps have the ability to identify the potential terms with high confidence. However, the figure also shows that the outcome of $NN(x,t;\theta)$ still has some deficiencies. For example, certain deviations will occur when faced with sparse data with high levels of noise (see Fig. 8(a)), and oscillations will emerge near the shock wave (see Fig. 8(b)). This limitation will bring a certain error to derivative calculation, which finally leads to the failure of PDE discovery when faced with complex situations, and this is why the PINN steps are utilized here. The figure reveals that after adding the potential terms discovered by the DLGA steps as physical constraints, the trained PINN is closer to the true data even with high noise and shock waves, which maintains the accuracy of derivative calculation and improves the performance of PDE discovery.

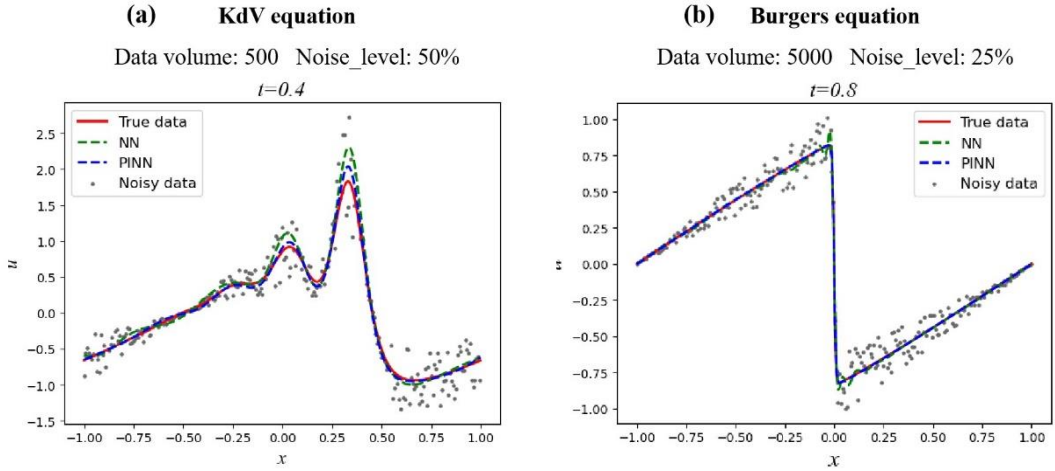

**FIG. 8.** True data, noisy data, and the outcome of $NN(x,t;\theta)$ and $PINN(x,t;\theta)$ for the KdV equation at $t$=0.4 (a) and the Burgers equation with a shock wave $t$=0.8 (b). Here, noisy data are plotted on each point of $x$ to better illustrate the data noise, and the utilized noisy data are actually sparser.

## IV CONCLUSION AND DISCUSSION

In this work, a novel framework, called the robust deep-learning genetic algorithm (R-DLGA), is proposed to handle complex situations, including sparse data with high noise, high-order derivatives and shock waves, which is difficult to deal with by existing methods. In the framework, a neural network is employed to learn a substitute model from the sparse and noisy data, and a new genetic algorithm, called the generalized genetic algorithm, is proposed to give a preliminary result of potential terms that is then added into the loss function of the physics-informed neural network (PINN) as physical constraints. The PINN is employed to further optimize the structure of the neural network to get closer to the actual physical processes and improve the accuracy of derivative calculation. A threshold is utilized to drop error compensation terms identified in the potential terms during the optimization process, and the reserved terms and their corresponding coefficients are the ultimate discovered PDE. The KdV equation, the KS equation, and the Burgers equation with a shock wave are employed to test our proposed algorithm. The results demonstrate that R-DLGA is stable and accurate in complex situations, including sparse data with high noise, high-order derivatives, and shock waves.

In the field of PDE discovery, the main concern is derivative calculation, which directly determines whether the correct PDE can be identified and the accuracy of the discovered PDE. However, faced with complex situations mentioned above, derivatives are difficult to be calculated accurately enough to discover the correct PDE, even if automatic differentiation is employed. Our proposed algorithm, however, finds another approach to improve the accuracy and stability of derivative calculation, which combines DLGA and PINN. Numerical experiments have shown that the neural network $NN(x,t;\theta)$ is able to learn the underlying dynamics to a certain extent, but is not sufficiently precise to directly discover the correct PDE in complex situations. Therefore, our proposed generalized genetic algorithm, which involves an inner term and a derivative order in the gene module, is utilized to discover potential terms. Different from the traditional genetic algorithm, the generalized genetic algorithm aims to discover a compound form of PDE terms, which is more concise and flexible. Therefore, the same term will have several different equivalent forms, which can increase the stability of the algorithm, and additional details about the influence of the equivalent forms are provided in Appendix A. Meanwhile, order mutation enables the generalized genetic algorithm to automatically add high-order derivatives if low-order derivatives are not sufficient. It is worth noting that values of PDE terms in compound form are calculated via the finite difference method, which means that high-order derivatives do not need to be previously defined and calculated since they can be calculated automatically based on several basic genes. Consequently, our proposed algorithm can discover PDEs in a wide range of variation library and solve the problem of an incomplete candidate library. Results have also demonstrated that the generalized genetic algorithm is able to identify the correct PDE terms with relatively accurate coefficients. Furthermore, several high-order error compensation terms with tiny coefficients are found, which means that the generalized genetic algorithm is stable to contain the correct terms in the potential terms, and it will pave the way for the PINN steps.

In this work, numerical experiments are carried out to examine the effect of the PINN steps, and the results have proven that the physical constraints provided by potential terms can significantly increase the accuracy of derivatives calculation, since it is closer to the underlying physical process even with shock waves and high noise. Different from the DeepMod method proposed by Both et al. [24], in which a pre-determined complete candidate library with numerous terms is defined as physical constraints, our proposed algorithm employed the generalized genetic algorithm to identify a few potential terms as physical constraints. This makes our optimization cycle more efficient, and the selection of threshold $\lambda$ in the PINN is not crucial because it is easier to maintain sparsity since terms in the physical constraints are parsimonious. In addition, different PDE discovery methods are adopted to discover the KdV equation from sparse data with high noise. The result shows that existing methods still possess certain defects since their performances are unsatisfactory, while R-DLGA succeeds to discover the correct PDEs with high accuracy.

Overall, our proposed R-DLGA algorithm is robust and accurate in various complex conditions, such as sparse data with noisy data, high-order derivatives, and shock waves. Moreover, it can discover PDEs from a large variation library and solves the problem of an incomplete candidate library. Despite these advantages, our proposed algorithm still possesses certain limitations and challenges. Firstly, although our proposed method is able to discover PDEs stably and accurately, the current research is still in the stage of proof-of-concept. Therefore, applications in practical problems require further investigation. Secondly, only one-dimensional

PDEs are investigated. Since discovery of PDEs with higher dimension in complex situations is more challenging, further improvement of R-DLGA to suit high dimensional PDEs is needed.

**Acknowledgements**

This work is partially funded by the Shenzhen Key Laboratory of Natural Gas Hydrates (Grant No. ZDSYS20200421111201738) and the SUSTech --- Qingdao New Energy Technology Research Institute.

## APPENDIX A: EQUIVALENT FORM OF TERMS IN THE GENERALIZED GENETIC ALGORITHM

In the proposed generalized genetic algorithm, terms are expressed in compound form, which comprises the inner term and the derivative order. Therefore, the same term will have several different equivalent forms. For example, the term $uu_x$ may be expressed as [(0,0):1] that refers to $(u^2)_x$, or [(0,1):0] that refers to $uu_x$, and both of them are equivalent. In this part, the influence of the equivalent form of terms in the genetic algorithm is investigated, and the KdV equation is taken as an example. 25,000 data are randomly selected to generate the new dataset, and 5% noise is added to the dataset. The correct PDE has several equivalent forms, and four of them are selected as examples. The fitness of different equivalent forms of the correct PDE and corresponding coefficients calculated in the genetic algorithm are presented in Table AI. From the table, it can be seen that the fitness of the equivalent forms and corresponding coefficients are slightly different. The common genetic algorithm can only discover the form in the first row ($u_{xxx}$ and $uu_x$); however, our proposed generalized genetic algorithm is able to discover an equivalent form with smaller (better) fitness, which is more accurate and stable.

**TABLE AI.** The fitness of four equivalent forms of the KdV equation and corresponding coefficients calculated in the genetic algorithm.

| Genome | Equation | Fitness |
|---|---|---|
| {[(0):3],[(0,1):0]} | $u_t = -0.002424 u_{xxx} - 0.9702 uu_x$ | 0.4079 |
| {[(0):3],[(0,0):1]} | $u_t = -0.002412 (u)_{xxx} - 0.4854 (u^2)_x$ | 0.4085 |
| {[(3):0],[(0,1):0]} | $u_t = -0.002415 u_{xxx} - 0.9699 uu_x$ | 0.4092 |
| {[(3):0],[(0,0):1]} | $u_t = -0.002421 u_{xxx} - 0.4856 (u^2)_x$ | 0.4068 |